\documentclass[letterpaper, 10 pt, journal, twoside]{IEEEtran}

\usepackage{cite}
\usepackage{multirow}
\usepackage{amsmath,amssymb,amsfonts}
\usepackage[pdftex]{graphicx}
\usepackage[caption=false,font=footnotesize]{subfig}
\usepackage{textcomp}
\usepackage{xcolor}
\usepackage{tikz}
\usepackage{booktabs}
\usepackage{adjustbox}
\usepackage{rotating}
\usepackage{layouts}
\usepackage{float}
\usepackage{gensymb}
\usepackage{dblfloatfix}

\IEEEoverridecommandlockouts                              % This command is only needed if 
\title{\LARGE \bf
SCCRUB: Surface Cleaning Compliant Robot Utilizing Bristles
}

\author{Jakub F. Kowalewski,$^{1}$ Keeyon Hajjafar,$^{2}$ Alyssa Ugent$^{1}$ and Jeffrey Ian Lipton$^{1}$% <-this % stops a space
\thanks{$^{1}$Jakub Kowalewski, Alyssa Ugent, and Jeffrey Lipton are with the Department of Mechanical and Industrial Engineering, Northeastern University, Boston, MA 02310, USA
        {\tt\small kowalewski.j@northeastern.edu, alys4321c@gmail.com, j.lipton@northeastern.edu}}%
\thanks{$^{2}$Keeyon Hajjafar is with the Department of Computer Science, Northeastern University,
        Boston, MA 02130, USA
        {\tt\small hajjafar.k@northeastern.edu}}%
}

\begin{document}

\maketitle
\thispagestyle{empty}
\pagestyle{empty}

%%%%%%%%%%%%%%%%%%%%%%%%%%%%%%%%%%%%%%%%%%%%%%%%%%%%%%%%%%%%%%%%%%%%%%%%%%%%%%%%
\begin{abstract}

%Basic Introduction
Scrubbing surfaces is a physically demanding and time-intensive task.
%Detailed background
Removing adhered contamination requires substantial friction generated through pressure and torque or high lateral forces.
%General Problem
Rigid robotic manipulators, while capable of exerting these forces, are usually confined to structured environments isolated from humans due to safety risks. In contrast, soft robot arms can safely work around humans and adapt to environmental uncertainty, but typically struggle to transmit the continuous torques or lateral forces necessary for scrubbing.    
%Here We show
Here, we demonstrate a soft robotic arm scrubbing adhered residues using torque and pressure, a task traditionally challenging for soft robots.
%Main Result
We train a neural network to learn the arm's inverse kinematics and elasticity, which enables open-loop force and position control. Using this learned model, the robot successfully scrubbed burnt food residue from a plate and sticky fruit preserve from a toilet seat, removing an average of 99.7\% of contamination.
%General Context
This work demonstrates how soft robots, capable of exerting continuous torque, can effectively and safely scrub challenging contamination from surfaces.
\end{abstract}

%%%%%%%%%%%%%%%%%%%%%%%%%%%%%%%%%%%%%%%%%%%%%%%%%%%%%%%%%%%%%%%%%%%%%%%%%%%%%%%%
\section{INTRODUCTION} 

Cleaning is a labor-intensive task that exposes workers to various health risks \cite{lin2022safety}. The most difficult-to-remove contamination typically requires scrubbing, which consists of high forces and repetitive motions that carry an increased risk of musculoskeletal injury \cite{Yassi1997,Woods2005-po}. Demand for cleaning robots is growing, with a global market projected to reach 40.8 billion by 2032 \cite{marketus_cleaning_robot_2024}. Within this market, scrubbing robots have primarily been deployed in the outdoors to clean boat hulls \cite{Souto2013-nj}, swimming pools \cite{Simoncelli2000-zc}, windows \cite{nam2019gas}, and solar panels \cite{Patil2017-hx}. However, in human-occupied environments, robotic scrubbers remain largely confined to cleaning floors \cite{forlizzi2006service}. There exists an unmet demand for robots that can safely work around humans and scrub common household surfaces such as plates, tables, and toilets.

\begin{figure}[ht]
    \centering
    %\hfill
    \includegraphics[width=1\linewidth]{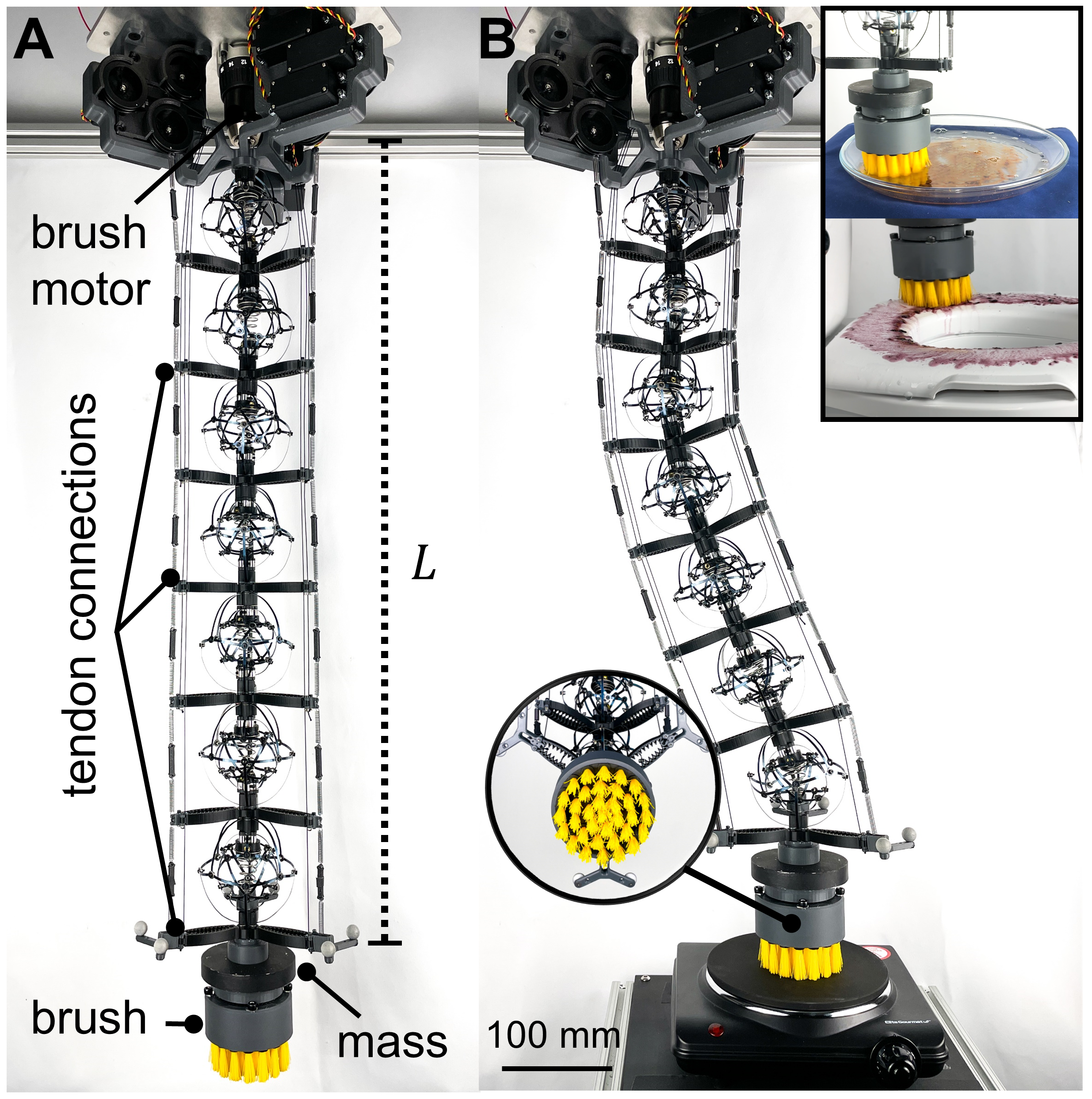}
    \caption{Overview of the soft scrubbing robot. (A) The arm ($L=710$ mm) is actuated using tendons, and a brush and mass are mounted at the end effector. A motor at the base drives the brush by transmitting torque concentrically through the arm's body. (B) The arm's compliance enables safe high-contact surface interactions.}
    \label{fig:FBD}
\end{figure}

% Background on scrubbing.
Scrubbing relies on generating frictional forces to break the adhesion bonds between contaminants and a surface \cite{Xu2005-dt}. Unlike loose particles or liquids that can be swept or wiped with minimal contact, adhered contamination is typically removed using substantial frictional forces from a tool such as a brush, sponge, or cloth \cite{johansson2007handbook}. These frictional forces can be generated at the surface-tool interface by exerting normal forces and torque or large lateral forces \cite{Huang2011-yt}. In particular,  spinning brushes are highly effective at removing surface contamination \cite{Langoyli-Giske2020-az}.

% Rotary tools and robots
Robot manipulators that are fully rigid can exert large forces and torques. However, their high stiffness and inertia pose safety risks \cite{Braganca2019-ur}, often leading them to work in structured and isolated environments. In contrast, soft robotic arms that compliantly deform on contact can safely operate around humans \cite{Guan2023-gj}. However, soft manipulators typically struggle to provide the continuous torque or sustained lateral forces required for effective scrubbing.

% Here we show
In this paper, we introduce SCRUBB, a Surface Cleaning Compliant Robot Utilizing Bristles. This work builds on the TRUNC arm, which transmits continuous torque using soft extensible, constant-velocity couplings \cite{carton2025bridging}. We designed a counter-rotating brush that reduced the reaction moment on the arm when scrubbing by $85\%$, compared to a standard brush. We then extended our prior modeling framework to account for deflections under large contact forces. We collected a force-deflection dataset and trained a neural network for trajectory planning, which achieved a mean error of 7.00 mm (0.98\% of the arm's length) and $1.3^{\circ}$. We also demonstrated open-loop force control, allowing us to modulate the brush's contact pressure between 1 to 8.5 N. Finally, we used the robot to scrub burnt ketchup from a plate and fruit preserve from a toilet seat. The robot successfully removed 99.7\% of the contamination on average compared to 47.1\% with wetting and rinsing alone. To summarize, we:
\begin{enumerate}
\item Demonstrated a soft robot arm capable of effectively scrubbing and cleaning a plate and toilet seat.
\item Expanded the control framework for soft torque transmitting arms to be able to vary forces applied to surfaces.
\item Designed an end effector that minimizes the reaction moment during scrubbing. 

\end{enumerate}

\section{RELATED WORK} 

\begin{figure}
    \centering
    %\hfill
    \includegraphics[width=1\linewidth]{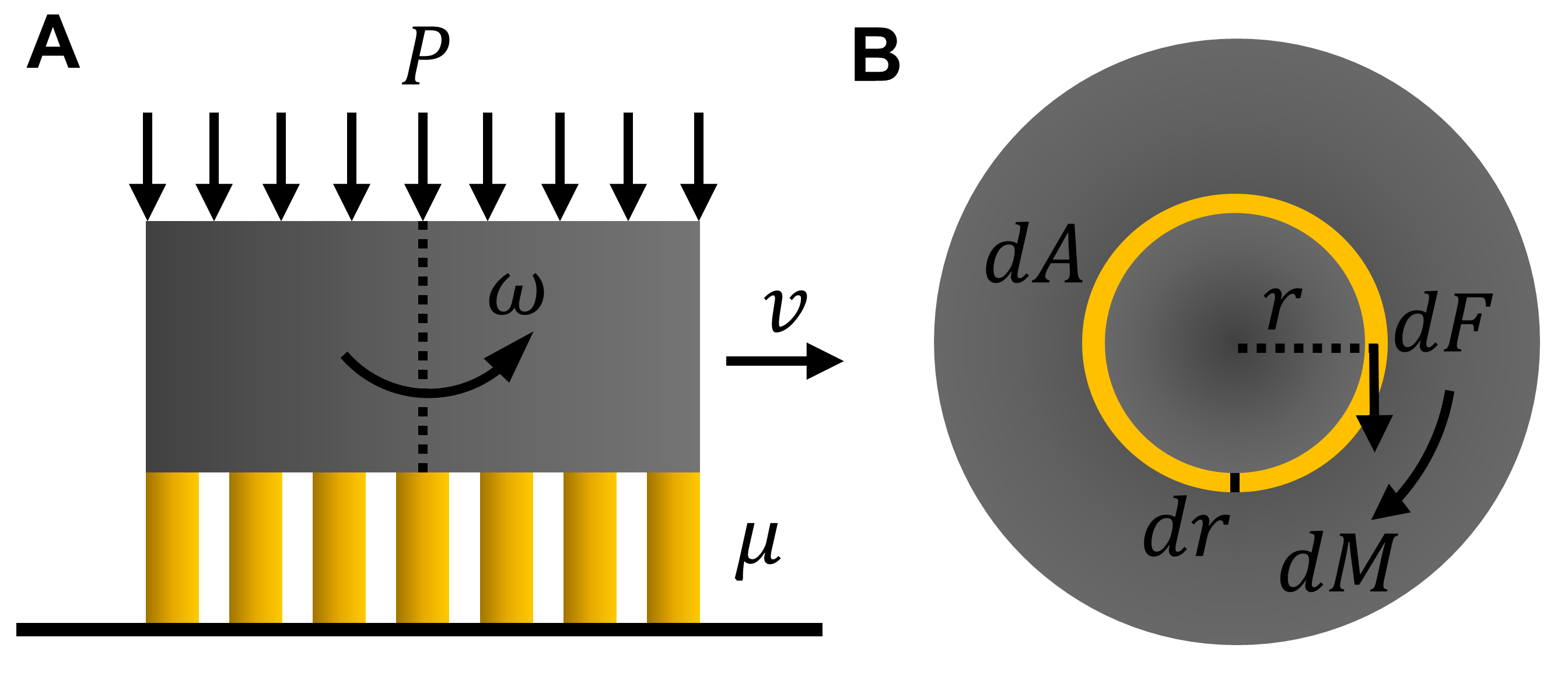}
    \caption{(A) We model scrubbing as a brush under uniform pressure $P$, that spins with an angular velocity $\omega$, and translates with a speed $v$. (B) For low translational speeds, we model dynamic friction between the bristles and surface as a net moment acting on the brush.}
    \label{fig:brush-diagram}
\end{figure} 

% 1) Rigid manipulators
Kitchens and bathrooms require some of the most frequent cleaning in the home \cite{scott1982investigation}. For robots to clean these spaces effectively, they must be able to scrub a variety of surfaces such as sinks, toilets, dishware, counters, and stovetops. Humanoid robots \cite{Kim2018-yv,Leidner2019-gm,Dometios2018-lh} and mobile manipulators \cite{Fu2024-kg,Lew2023-yw} have been used to wipe loose debris and liquids from tables and counters using cloths and sponges. However, wiping alone is insufficient for removing adhered contamination. Wakabayashi et al. demonstrated more forceful contact using a humanoid robot to scrub dishes with a sponge \cite{Wakabayashi2024-sy}, but maintaining sufficient scrubbing pressure while ensuring safety remains a challenge for fully rigid systems.

% 2) Compliant end effectors 
To improve force control during scrubbing, some systems introduce mechanical compliance at the end effector. Harmatz et al. combined a rigid arm with series elastic actuators (SEAs) to enable safer scrubbing and better force control \cite{harmatz2024hybrid}. Suh et al. developed compliant bubble grippers with force feedback and used a squeegee to wipe liquid from a table \cite{Suh2022-pk}. Huang et al. used a soft tactile-sensing end-effector to wipe contoured surfaces, such as curved pipes and bookshelves  \cite{Huang2022-ek}. While SEAs can improve contact safety between a tool and surface, the rigid arms they attach to still pose safety risks around humans.

% 3) Fully compliant systems
Soft robotic arms, made from compliant materials and structures, offer inherent safety in unstructured environments and around people. This compliance enables them to safely wipe humans and assist with bathing tasks \cite{Zlatintsi2020-jk, Manti2016-th}. Tang et al. demonstrated simultaneous control of position and normal force while wiping surfaces with a soft arm \cite{Tang2022-xz}. Although soft arms can generate substantial normal forces \cite{Alessi2024-tn}, they typically struggle to exert large lateral forces, making scrubbing challenging. Rotational scrubbing is an alternative approach that generates friction through spinning rather than lateral motion. However, most soft arms are unable to transmit continuous torque. 

% 4) How are we different
Our work differs from these by enabling a soft robot to generate frictional scrubbing forces using torque and pressure rather than lateral forces. This is possible due to the arm's high twist-to-bend ratio that allows it to spin a brush while remaining flexible under contact. This approach combines the adaptability and safety of soft structures with the torque transmission capabilities of rigid systems. We demonstrate that our soft arm extends beyond low-contact wiping tasks, effectively scrubbing tough contamination adhered to surfaces.  

\begin{figure}[t]
    \centering
    %\hfill
    \includegraphics[width=0.74\linewidth]{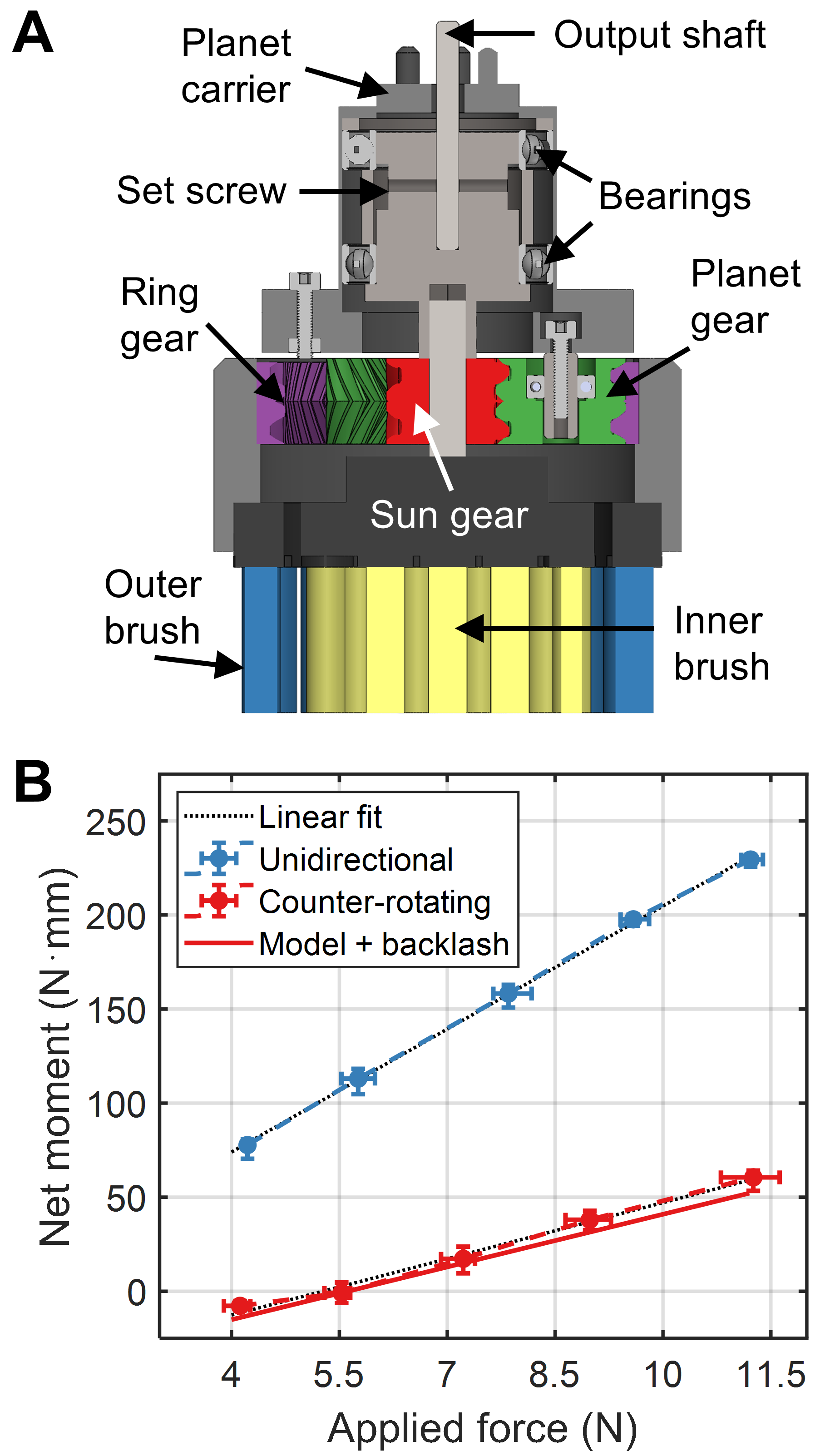}
    \caption{(A) Design of the counter-rotating brush based on a planetary gearbox. (B) We found the net torque produced by the counter-rotating brush was substantially lower than the unidirectional brush. Error bars represent the min and max values from $n=5$ trials.}
    \label{fig:Brush}
\end{figure} 

\section{Modeling Scrubbing} \label{sec:scrub_model}

We model the friction from a spinning brush with radius $r$ as a function of its angular velocity $\omega$, linear velocity $v$, applied pressure $P$, and dynamic coefficient of friction $\mu$ (Fig.~\ref{fig:brush-diagram}A). We assume for a low translational velocity, friction is dominated by $\omega$ and is tangent to the direction of spinning. This simplifies the frictional interaction to a net moment \( M \) at the brush-surface interface. Additionally, we assume a dense distribution of bristles and model the contact area as a continuous disc. To calculate the net moment, we define an infinitesimal ring of radius \(r\) and thickness \(dr\) (Fig.~\ref{fig:brush-diagram}B):

\begin{figure}[t]
    \centering
    %\hfill
    \includegraphics[width=0.9\linewidth]{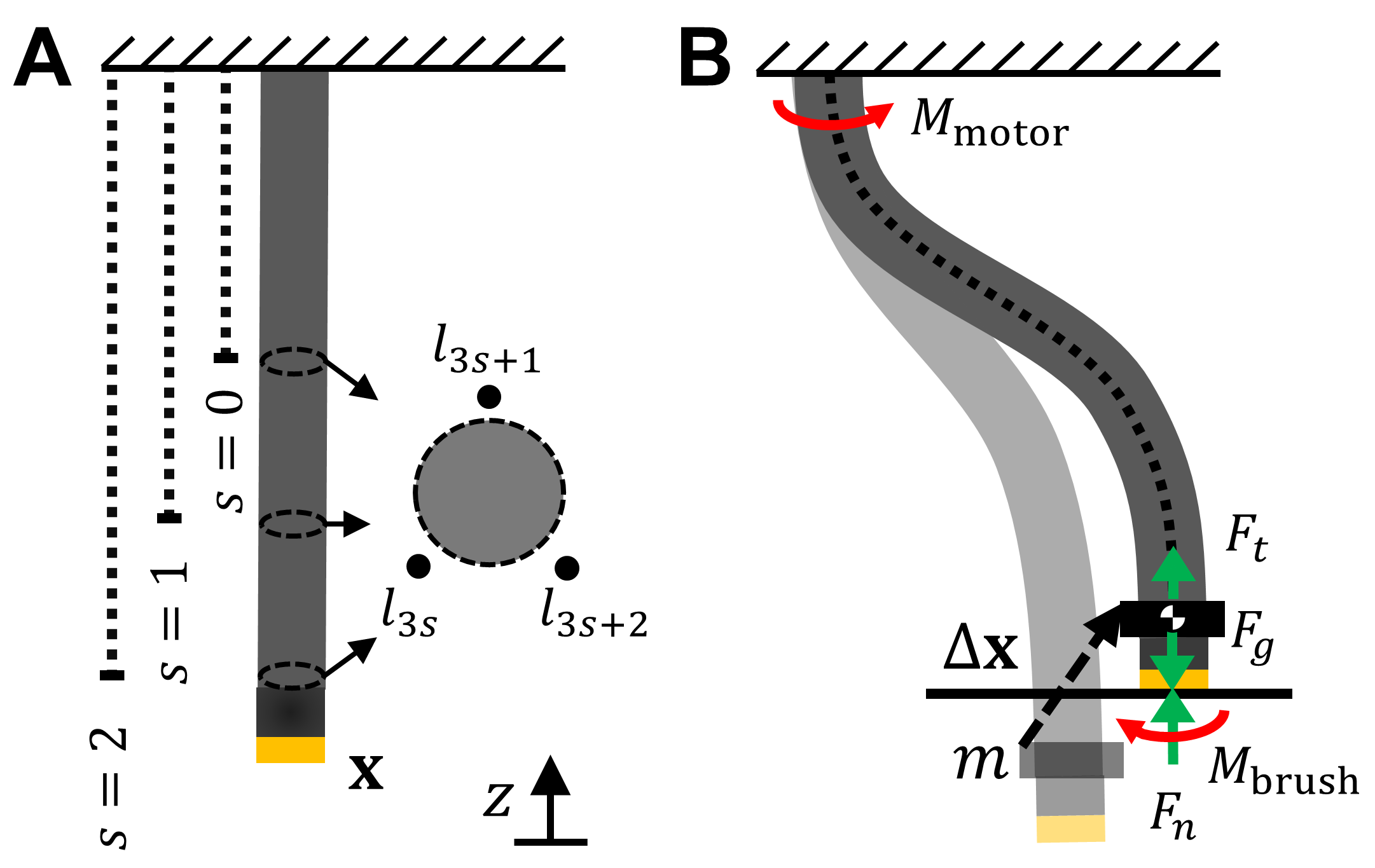}
    \caption{(A) The arm's configuration space is defined using the nine tendon lengths $l_i$, which connect to the arm at three stages $s$. (B) We assume the brush is heavier than the arm's body, and model it as a point load $F_g$ at the end effector.}
    \label{fig:fbd}
\end{figure} 

\begin{equation}
dA = 2\pi r\,dr
\end{equation}

The frictional force on this ring is:

\begin{equation}
dF = (\mu P)\, dA = \mu P \,(2\pi r\,dr)
\end{equation}

The force generates a differential moment $dM$:

\begin{equation}
dM = r\, dF = r\, (\mu P\, 2\pi r\,dr) = 2\pi \mu P\, r^2\,dr \label{eq:d_moment}
\end{equation}

Integrating Equation~\ref{eq:d_moment} over the brush's radius yields net moment:

\begin{equation}
    M = \int_0^{r} 2\pi \mu P\, r^2\,dr \label{eq:integral}
\end{equation}

In prior work, we transmitted torque while anchoring to the environment, such as when fastening bolts or turning valves. However, scrubbing with a brush introduces lateral degrees of freedom at the end effector that are difficult to control. When the brush is directly beneath the arm, the moments from the brush and motor are concentric, and the arm only experiences torsion. However, as the brush moves radially out, the moments are no longer concentric, and the arm experiences both torsion and bending. By design, the arm has a low flexural stiffness, which prevents it from effectively resisting bending. To stabilize the end effector, we use a counter-rotating brush with two faces that spin in opposite directions, minimizes the net torque.

The counter-rotating brush's inner face is a solid disc with radius $r_i$, while the outer face is an annulus with an inner radius of $r_i$ and outer radius $r_o$. Applying Equation~\ref{eq:d_moment}, the total moment for each brush face is:

\begin{align}
    M_{\text{inner}} &= \int_0^{r_i} 2\pi \mu P\, r^2\,dr 
= \frac{2\pi \mu P\, r_i^3}{3} \\
    M_{\text{outer}} &= \int_{r_i}^{r_o} 2\pi \mu P\, r^2\,dr
= \frac{2\pi \mu P}{3} \left(r_o^3 - r_i^3\right) \\
M_{net} &= M_{\text{outer}} - M_{\text{inner}} = \frac{2\pi\mu P}{3} (r_o^3-2r_i^3) \label{eq:net_moment}
\end{align}

We set the inner and outer brush moments equal to achieve a net-zero moment. We find the ratio of inner and outer radius to be $\frac{r_o}{r_i} = 2^{\frac{1}{3}} $
Thus, we find that for our scrubbing conditions, the inner radius of a counter-rotating brush should be $\approx79\%$ of the outer radius to achieve a net zero moment.

\begin{figure}
    \centering
    %\hfill
    \includegraphics[width=0.75\linewidth]{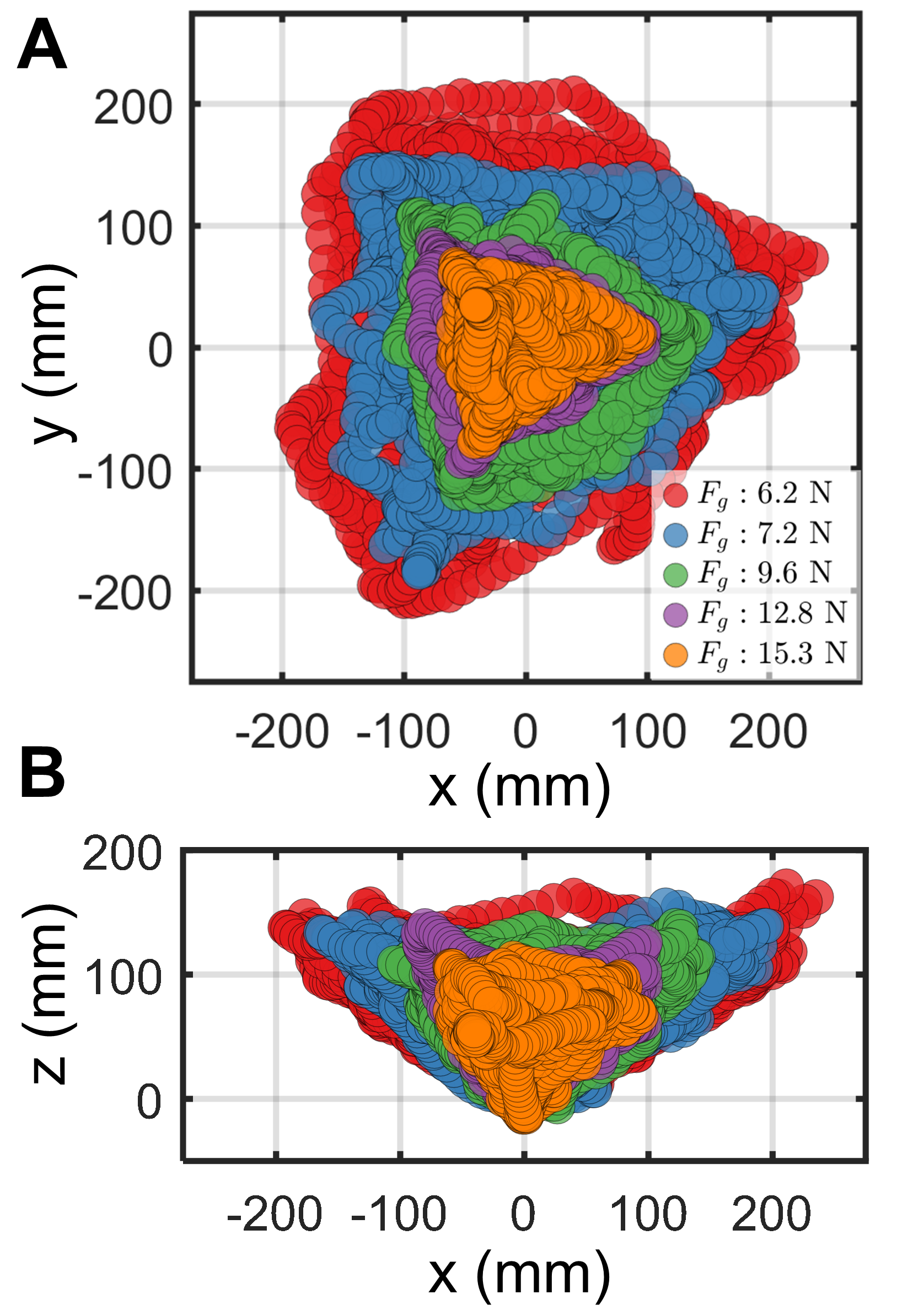}
    \caption{The force-deflection dataset (10,000 samples) used to train the neural network viewed from (A) the top and (B) the side. Each color represents a different loading condition for $F_g$ as indicated in the legend.}
    \label{fig:Workspace}
\end{figure} 

\begin{figure*}
    \centering
    %\hfill
    \includegraphics[width=1\linewidth]{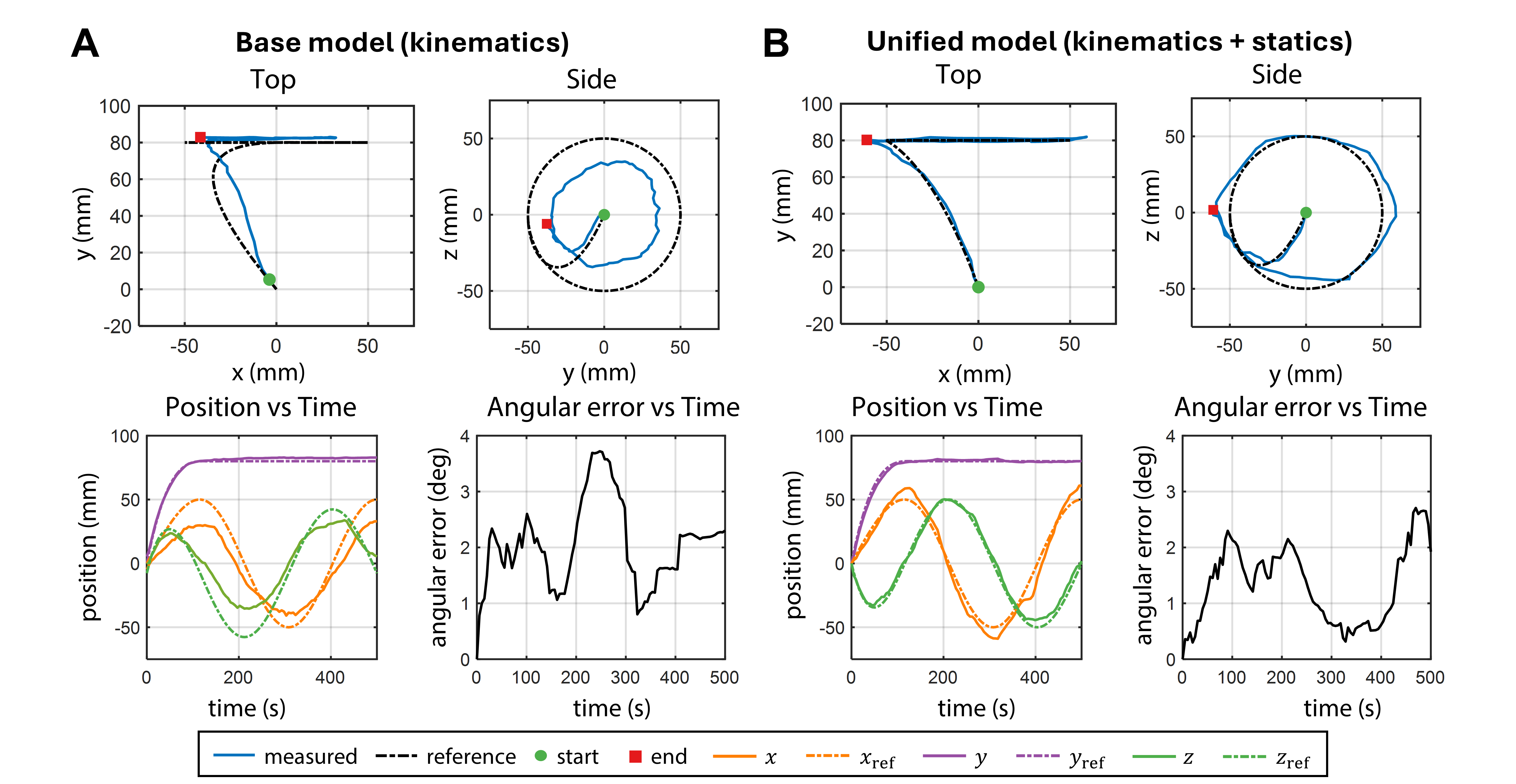}
    \caption{ Using the neural network model, we generated a circular trajectory for the arm to follow. We attached a $3.4$ N weight at the end effector to generate an external load. (A) Using our original network, the arm deflects inward and tracks a smaller circle. (B) The new model compensates for this deflection, allowing the arm to track the circle more accurately.}
    \label{fig:Trajectory}
\end{figure*}

\section{Scrubber Design} \label{sec:brush}

\subsection{Design and fabrication}

The counter-rotating brush we designed splits a single input rotation ($\omega=110$ RPM) into two opposite rotations using a planetary gearbox (Fig.~\ref{fig:Brush}A). We purchased a scrubbing brush (3.5 in diameter) and milled a circular slot using a CNC router to produce the inner and outer faces. The radii ($r_o = 35$ mm and $r_i = 25$ mm) were selected to match the ratio calculated in Section \ref{sec:scrub_model} as closely as possible. However, due to constraints of the radial bristle spacing, the brush's outer radius is 11.1\% larger than the ideal outer radius from our model. Based on this error, the model predicts the brush will generate $9.34$ N·mm of torque per newton of applied force.

The brush is driven by a motor at the base, which connects directly to the inner face. The sun gear (16 teeth) is coupled directly with the inner brush's hex shaft. The three planet gears (15 teeth) are fixed to the arm's outer shaft, which is rotationally grounded. Each planet gear has a bearing (McMaster part: 7804K1) mounted in its center. Since the planet carrier does not rotate, it serves as the mounting point for additional masses. The planet carrier also houses a set of bearings (McMaster part: 5972K216) that allow the coupler to spin freely. Finally, the ring gear (50 teeth) is coupled to the outer brush face using cyanoacrylate. We fabricated the gearbox parts using a Prusa MK4S 3D printer using Polylactic Acid (PLA) filament.

\subsection{Net-moment during scrubbing}

To validate our brush design, we measured the reaction moment as a function of normal force and compared it with a unidirectional brush. We collected the force and torque measurements using an Instron machine with a $\pm450$ N and $\pm5$ N·m biaxial load cell. We placed a sheet of 60-grit abrasive on top of the compression anvil to simulate surface contamination. To allow for the normal force to be gradually increased, we placed a spring in series with the brush. We then collected data from a pre-compression of 4 N to 11 N when the spring was fully compressed.

At each force increment, we completed $n=5$ trials (error bars represent min and max values) and averaged the moment data over four seconds (Fig.~\ref{fig:Brush}B). We then fit linear models to the data with $R^2$ values of 0.999 and 0.990 for the unidirectional and counter-rotating data, respectively. From the unidirectional brush data, we extracted the dynamic coefficient of friction $\mu$ using the relation $M = \frac{2}{3} \mu r F$, obtained from Equation~\ref{eq:integral}. Substituting the best-fit slope and known radius into the equation yielded an estimate of $\mu \approx 0.93$. We then used this coefficient to predict the slope of the moment–force relationship for the counter-rotating brush.

For a counter-rotating brush, the force-moment relationship is $M = \frac{2}{3} \mu (r_o - 2r_i) F$, based on Equation~\ref{eq:net_moment}. Plugging in our brush dimensions, we obtain a slope of 9.34 N·mm/N, which closely matches the slope of the experimental data's linear fit of 9.96 N·mm/N. However, the intercept for the counter-rotating data is shifted. We believe this is due to backlash in the gearbox, which requires some initial force before both brush faces fully engage, shifting the linear regime. To account for this, we introduce a backlash term $\delta$, to the model $M = kF - \delta$, where $k$ is the predicted slope. From our linear fit, we found $\delta = 52.4$ N·mm, which we incorporated into the model and show it closely matches our experimental data (Fig.~\ref{fig:Brush}B). Overall, we demonstrated that the counter-rotating brush reduced net moment by an average of 85\% compared to the unidirectional brush.

\section{Modeling the Arm} \label{sec:brush}

\subsection{Deflection under loading}

Scrubbing requires exerting a sustained forces while maintaining positional control over a tool. For soft manipulators, these external forces result in substantial deflection at the end effector. In this section, we extend our prior learning approach for trajectory planning \cite{carton2025bridging} to account for the arm's elastic behavior under external loading.

In the scope of this work, we consider cases where the brush is scrubbing a planar surface, normal to gravity. We begin by defining the arm's configuration space $\mathbf{q} \in \mathbb{R}^9$ using the nine tendon lengths $l_i$, which connect at three stages $s$ along the arm and follow the numbering convention in Figure.~\ref{fig:fbd}A. We then define the end effector's position and orientation as the vector $\mathbf{x} = [x,y,z,q_w,q_x,q_y,q_z]$, where the orientation is encoded as a quaternion. The external loads ($F_g$ and $F_n$) cause the arm to deflect by $\Delta \mathbf{x}$ (Fig.~\ref{fig:fbd}B), shifting the inverse-kinematic mapping between $\mathbf{q}$ and $\mathbf{x}$. 

Under the assumption that the translational speed of the brush is low, we can express the arm's loading during scrubbing as quasi-static equilibrium:

\begin{align}
    \sum F_x &= 0 \\
    \sum F_y &= 0 \\
    \sum F_z &= F_t + F_n - F_g = 0 \label{eq:Fz}
\end{align}

From Equation~\ref{eq:Fz}, we can solve for the arm's internal tendon tension \( F_t \) since $F_g$ is known and $F_n$ is a control input:

\begin{equation}
    F_t = F_g - F_n
    \label{F_arm}
\end{equation}

Finally, we define an augmented inverse kinematic mapping $f^{-1}$ that encodes the arm's kinematics and elasticity into a single function: 
\begin{equation}
    f^{-1}: \mathbb{R}^8 \rightarrow \mathbb{R}^9,\quad (\mathbf{x}, F_t) \mapsto \mathbf{q} \label{eq:mapping}
\end{equation}
Using this mapping, we can control the end effector's position and orientation while compensating for deflections due to normal loads.

\subsection{Learning the model}

We learned the mapping from Equation~\ref{eq:mapping} using a neural network. To train the model, we collected a force-deflection dataset of arm poses under various loading conditions. We varied the applied load by starting with the arm's base weight $F_g = 6.2$ N and incrementally adding mass until $F_g = 15.3$ N (Fig.~\ref{fig:Workspace}). For each weight, we sampled 2,000 poses, generating a total dataset of 10,000 pose-load combinations. During data collection, we used a feedback loop from the motion capture system to remove all cable slack, ensuring the sampled data was from the well-defined manifold of the configuration space. 

We randomly split the dataset into training (80\%) and validation (20\%) subsets. The model was implemented in PyTorch as a fully connected neural network with two hidden layers, each containing 1600 neurons and ReLU activations. We trained the network using the Adam optimizer with a learning rate of 0.001 and momentum of 0.9. We used a batch size of 16 and trained for 50 epochs. The loss function was the mean squared error between the predicted and actual tendon lengths. To improve convergence, we applied exponential learning rate annealing with a decay factor $\gamma=0.9$, which progressively reduced the learning rate after each epoch. The model was trained using an Nvidia GeForce RTX 4090 graphics processing unit (GPU), which took 42.1 seconds.  

We experimentally validated the model by programming the arm to follow a circular trajectory (100 waypoints with 0.5-second pauses) while hanging a $3.4$ N weight from the brush. We performed all inference on a 12th-generation Intel i7-12700 central processing unit (CPU), which took on average 1.33 ms per waypoint. We compared our new model to the neural network we previously trained on 18,300 poses with $F_g=6.2$. Using the previous model, we recorded a mean positional error of $15.7$ mm ($2.21\%L$) and a mean orientation error of $2.1^\circ$. The new model achieved mean positional and orientation errors of $7.0$ mm ($0.98\%L$) and $1.3^\circ$, respectively. These results demonstrate that the new model successfully compensates for external normal loads, enabling trajectory planning for scrubbing tasks.  

\subsection{Open-loop force control}

\begin{figure}
    \centering
    %\hfill
    \includegraphics[width=0.7\linewidth]{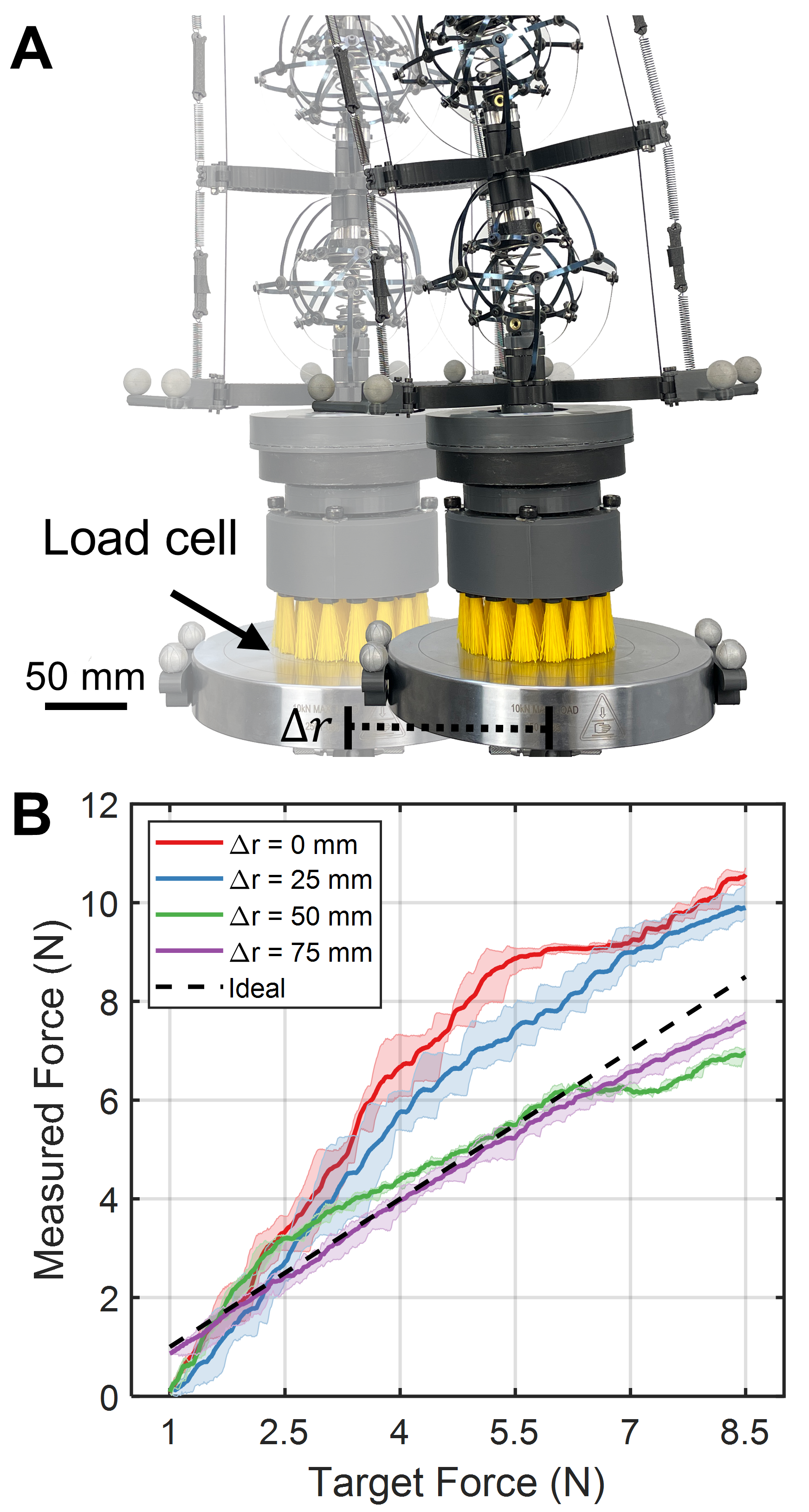}
    \caption{(A) We demonstrated open-loop force control over a sweep of radial distances. (B) We showed the arm can modulate the contact force from 1 N to 8.5 N, closely following the target linear ramp. Shaded error bars represent min and max from $n=5$ trials.}
    \label{fig:Force}
\end{figure} 

\begin{figure*}
    \centering
    %\hfill
    \includegraphics[width=1\linewidth]{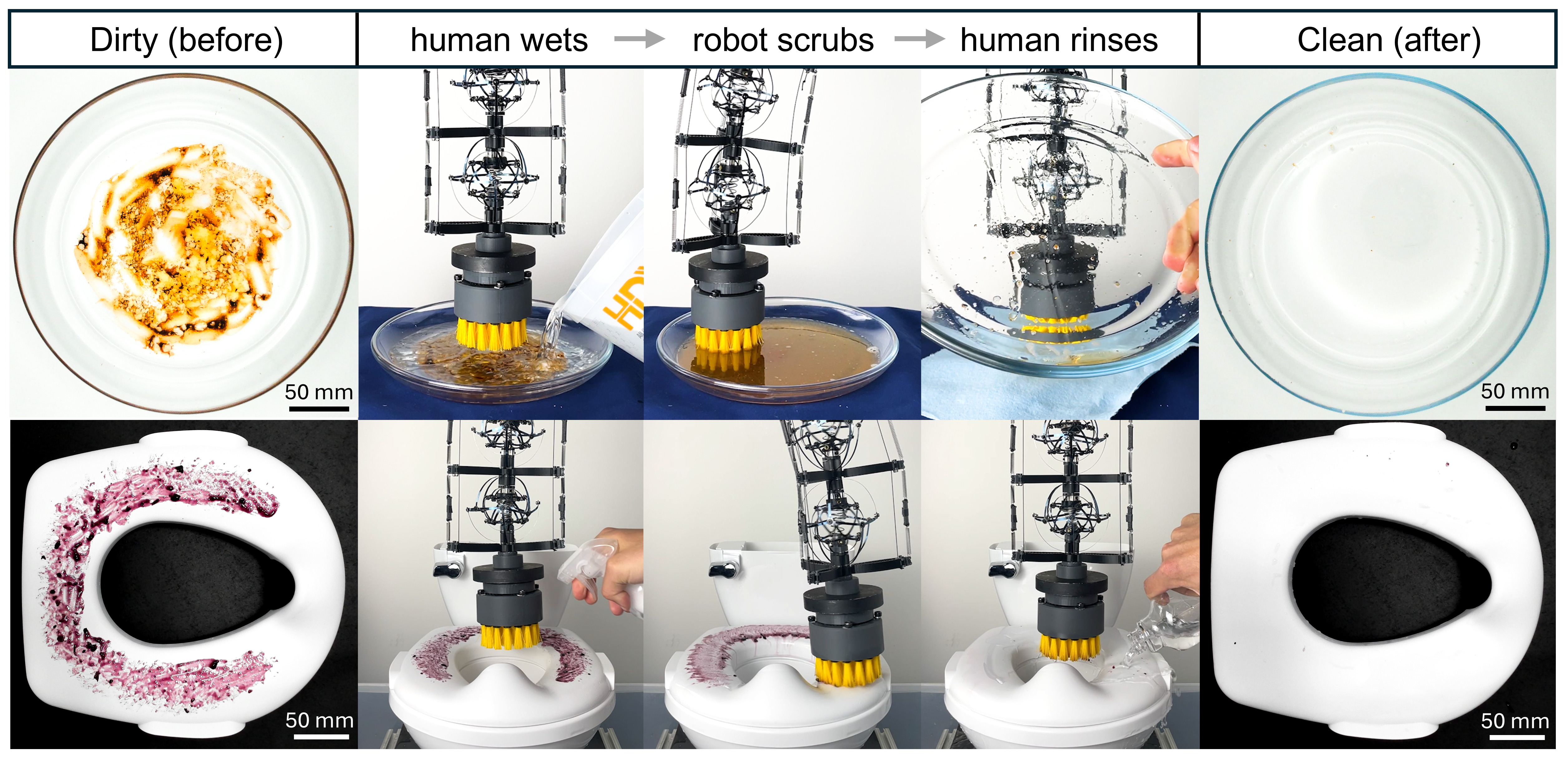}
    \caption{We used the arm to scrub a glass dish and toilet seat. A human wet the surfaces beforehand and rinsed once the arm finished. We simulated contamination using burnt ketchup and fruit preserve for the dish and the seat, respectively. The before and after images show that the brush effectively removed the contamination.}
    \label{fig:Demos}
\end{figure*}

Effective scrubbing requires applying and maintaining pressure on a surface. Here, we demonstrate using the neural network for open-loop force control. To exert a desired normal $F_n$ force at some position and orientation $\mathbf{x}$, we calculate $F_t$ using Equation~\ref{F_arm}. We then run an inference pass through the neural network to obtain the corresponding motor inputs $\mathbf{q}$. The maximum force $F_n$ the arm can apply while remaining in the training distribution is $F_g-F_n=F_{g,\text{min}}$ where $F_{g,\text{min}}$ is the lowest weight used during data collection. To reach the upper bound of our training, we set the arm's weight to $F_g=15.3$, allowing us to test up to $F_n=8.5$ N.

We then programmed the arm to apply a linear ramped force, starting with a minimal force of 1 N to ensure contact was made with the load cell. Since the arm's stiffness varies with position, we evaluated the force control over a radial sweep of $\Delta r$ (Fig.~\ref{fig:Force}A) from 0 to 75 mm, spanning the full workspace at $F_g=15.3$. For each trial, we fit a linear function $F_{\rm measured}=K_{\Delta r}\,F_{\rm target}+b_{\Delta r}$ with the ideal case being $K=1$, $b=0$. In order of increasing $\Delta r$, our $R^2$ values were 0.923, 0.965, 0.926, and 0.995. We found the gain errors, $(K_{\Delta r}-1)\times100\%$, to be 35.7\%, 35.0\%, 21.4\%, and 8.85\%, and offsets to be 0.23, -0.44, 0.90, and 0.17. Additionally, we found that when the motor spins, the contact force fluctuates with a standard deviation of 0.43 N. These results demonstrate that using the neural network and statics model, we can perform open-loop force control on planar surfaces.

\section{Scrubbing Tough Messes}

% Methods for cleaning the plate
We demonstrate the robot's scrubbing capabilities on two household surfaces: a glass plate and a pediatric toilet seat (Fig.~\ref{fig:Demos}).  In both tests, we set the system weight to $F_g = 9.6$ N and the target contact force to $F_n=3$ N, which we found produced sufficient pressure to remove dried food residue.

\subsection{Glass Plate}
To simulate kitchen contamination, we spread ketchup on the plate (288 mm diameter) and microwaved it for 90 seconds in a 950 W microwave until the ketchup began burning onto the surface. We then placed the contaminated plate underneath the robot arm and added soapy water. We programmed the arm to follow a circular trajectory (150 mm diameter), such that the edge of the brush closely followed the outer wall of the plate. The plate's vertical position relative to the arm was measured using the mocap to generate the path. After two passes around the perimeter, the arm moved to the center and dwelled for 5 seconds before retracting up, finishing in a total of 2 minutes and 45 seconds. Finally, we removed the plate, disposed of the dirty water, and rinsed it with 500 mL of clean water.

\subsection{Toilet seat}
For the toilet seat (290 mm length and 270 mm width), we simulated surface contamination using a blueberry fruit preserve, which naturally adhered to the surface. The scrubbing trajectory was then programmed by measuring a path along the middle of the seat using motion capture markers and generating a perimeter around the curve with a 27.5 mm offset towards the outer rim and 16 mm towards the inner rim to ensure full surface coverage. The seat was sprayed using soapy water before the arm began scrubbing. The arm completed its scrubbing pass in 1 minute and 49 seconds, and the seat was rinsed with $500$ mL of water at the end.

\subsection{Cleaning comparison}

\begin{figure}[h!]
    \centering
    %\hfill
    \includegraphics[width=.9\linewidth]{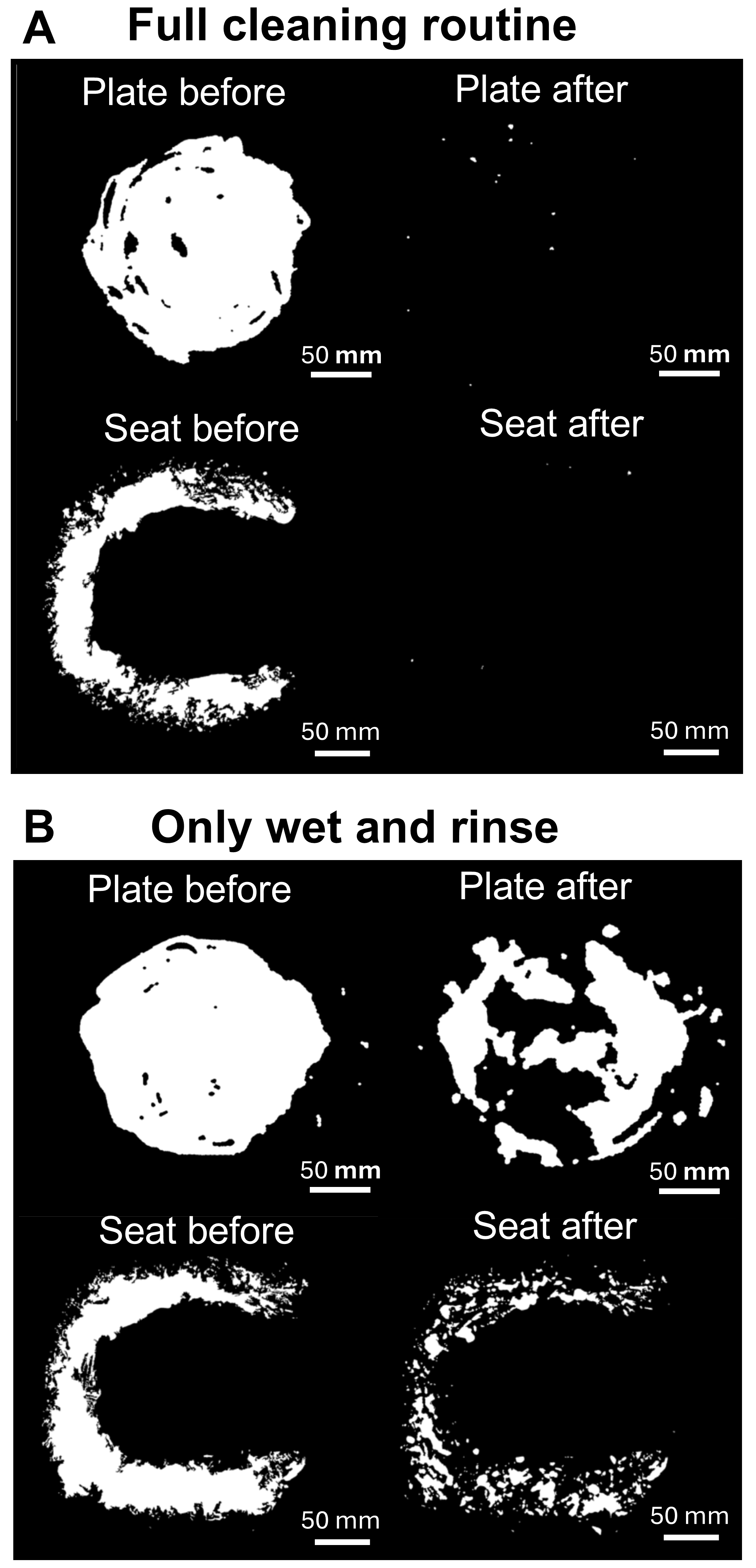}
    \caption{We compared (A) the robot's cleaning performance with (B) only wetting and rinsing the surfaces. The results demonstrate that the scrubbing robot was effective in removing stubborn contamination.}
    \label{fig:comparison}
\end{figure} 

We compared the arm's full scrubbing routine with only soaking and rinsing to quantify the cleaning effectiveness (Fig.~\ref{fig:comparison}). We measured the amount of contamination by segmenting the residues from the before and after RGB images. For the plate, we determined the region of interest (ROI) by detecting the circular rim using Canny edge detection and fitting a minimal enclosing circle to the largest contour. We radially padded this region by 30 pixels to exclude the red hue transmitted through the glass to the plate's edge. Within this ROI, we segmented the ketchup by selecting pixels where the red channel intensity exceeded 25 and was also greater than both the green and blue channels.

For the toilet seat, we first segmented the geometry by converting the image to grayscale and thresholding for pixel intensity above 200. The two largest contours were used to define the outer and inner boundaries of the seat. To exclude shadows, we padded the inner and outer regions by 20 and 30 pixels, respectively. Finally, the dark contamination spots were segmented by selecting pixels with an intensity below 150.

% Discussion of results
We found that the arm cleaned $99.7\%$ of the contamination compared to $32\%$ when soaking and rinsing the plate alone. For the toilet seat, the arm cleaned $99.8\%$ of the contamination compared with $61.4\%$ for spraying and rinsing only. The total contamination areas before and after are reported in Table~\ref{table}. Overall, these results demonstrate that the arm can effectively scrub adhered contamination that wetting and rinsing alone cannot remove.

\begin{table}
\centering
\small % if you need to squeeze it a bit
\caption{Surface cleaning comparison.}
\begin{tabular}{@{}l l c c@{}}
\toprule
\textbf{Surface} & \textbf{Metric}      & \textbf{Wet + Scrub + Rinse} & \textbf{Wet + Rinse} \\
\midrule
\multirow{3}{*}{Plate}
 & Before (px)       & 575188 & 679327 \\
 & After (px)        &  1624 &  455273 \\
 & Cleaned      & \bf{99.7\%} & 32.9\% \\
\midrule
\multirow{3}{*}{Toilet}
 & Before (px)       & 425240 & 537547 \\
 & After (px)        &  450 &  207614 \\
 & Cleaned      & \bf{99.8\%} & 61.3\% \\
\bottomrule
\end{tabular}
\label{table}
\end{table}

\section{Conclusion and Future Work}

In this paper, we demonstrated a soft robot scrubbing difficult-to-remove contamination from household surfaces. We presented a quasi-static scrubbing model and extended our prior learning framework to compensate for deflections under loading and control contact forces at the end effector. We then collected a force-deflection dataset of 10,000 poses and trained a new neural network, achieving an average accuracy of 7.00 mm (0.98\%L) and $1.3^{\circ}$ in our circle tracking test. We then presented a brush design that reduces the reaction moment during scrubbing by $85\%$ compared to a regular brush. We also demonstrated open-loop force control, allowing us to modulate the brush's normal force from 1 to 8.5 N. Finally, we used the arm to scrub burnt ketchup from a plate and fruit preserve from a toilet seat. On average, the arm removed $99.7\%$ of contamination, demonstrating effective performance in both kitchen and bathroom scenarios.

Future work includes extending the contact-force model to handle non-planar and irregular geometries. Integrating air and water spray nozzles into the brush would enable the arm to wet, rinse, and dry surfaces autonomously. Incorporating a vision system could allow the robot to detect contamination, generate cleaning paths, and identify any missed areas. For tasks requiring more precise force control, such as sanding, closed-loop control could be achieved using an in-line load cell.

% \addtolength{\textheight}{-12cm}   % This command serves to balance the column lengths
%                                   % on the last page of the document manually. It shortens
%                                   % the textheight of the last page by a suitable amount.
%                                   % This command does not take effect until the next page
%                                   % so it should come on the page before the last. Make
%                                   % sure that you do not shorten the textheight too much.

%%%%%%%%%%%%%%%%%%%%%%%%%%%%%%%%%%%%%%%%%%%%%%%%%%%%%%%%%%%%%%%%%%%%%%%%%%%%%%%%

%%%%%%%%%%%%%%%%%%%%%%%%%%%%%%%%%%%%%%%%%%%%%%%%%%%%%%%%%%%%%%%%%%%%%%%%%%%%%%%%

%%%%%%%%%%%%%%%%%%%%%%%%%%%%%%%%%%%%%%%%%%%%%%%%%%%%%%%%%%%%%%%%%%%%%%%%%%%%%%%%

% \section*{ACKNOWLEDGMENT}

% \TK{To do}

%%%%%%%%%%%%%%%%%%%%%%%%%%%%%%%%%%%%%%%%%%%%%%%%%%%%%%%%%%%%%%%%%%%%%%%%%%%%%%%%

\bibliographystyle{IEEEtran}
\bibliography{IEEEabrv,references}

\end{document}